\documentclass[pmlr,twocolumn,a4paper,10pt]{jmlr}
\usepackage{isipta2021}
\usepackage[american]{babel}
\usepackage{booktabs} 
\usepackage{tikz,bm,pgfplots}
\usepackage{algorithm,algorithmic}
\usepackage{listings}

\title{CREPO: An Open Repository to Benchmark Credal Network Algorithms}
\author{\Name{Rafael Caba\~nas}\Email{rcabanas@idsia.ch}\\
\Name{Alessandro Antonucci}\Email{alessandro@idsia.ch}\\
\addr Istituto Dalle Molle di Studi sull'Intelligenza Artificiale (IDSIA), Switzerland}
\begin{document}
\maketitle
\begin{abstract}
Credal networks are a popular class of imprecise probabilistic graphical models obtained as a Bayesian network generalization based on, so-called \emph{credal}, sets of probability mass functions. A Java library called CREMA has been recently released to model, process and query credal networks. Despite the NP-hardness of the (exact) task, a number of algorithms is available to approximate credal network inferences. In this paper we present CREPO, an open repository of synthetic credal networks, provided together with the exact results of inference tasks on these models. A Python tool is also delivered to load these data and interact with CREMA, thus making extremely easy to evaluate and compare existing and novel inference algorithms. To demonstrate such benchmarking scheme, we propose an approximate heuristic to be used inside variable elimination schemes to keep a bound on the maximum number of vertices generated during the combination step. A CREPO-based validation against approximate procedures based on linearization and exact techniques performed in CREMA is finally discussed.
\end{abstract}
\section{Introduction}\label{sec:intro}
Probabilistic graphical models are popular tools for modern AI and machine learning \cite{koller2009}. Although mostly based on classical probability theory, graphical models relying on more general formalisms have been also proposed \cite{borgelt2000possibilistic,yaghlane2008inference}. Among them, \emph{credal networks} \cite{cozman2000credal} display a clear and interpretable semantics, directly extending that of Bayesian networks \cite{pearl2014probabilistic}. Moreover, credal networks have been recently used to address unidentifiable queries in structural causal models \cite{zaffalon2020a}. The counterpart of such interpretability and expressiveness is the intractability of exact inferences. Being a generalization of Bayesian network inference \cite{cooper1990}, credal network inference is NP-hard and, unlike the case of Bayesian networks, might remain NP-hard even for simpler topologies \cite{maua2014probabilistic}. Nevertheless, a number of approximate algorithms has been proposed in the literature (see \cite{maua-2020-ijar-b} for a recent survey), thus making possible to practically use these models in applications (e.g., \cite{antonucci2004a,antonucci2009credal,estrada2019probabilistic}). 

So far, the software implementations of these algorithms are mostly developed by different research groups and based on different tools and formats, this preventing extensive benchmarking and comparison. Yet, a Java library for credal network modelling, processing and inference, called CREMA, has been recently released \cite{huber2020a}. The main goal of CREMA is to provide a standard and unifying framework to develop, test and compare existing credal network algorithms and promote the development of novel schemes. To fully accomplish this plan, we present here CREPO, an open and extensible repository including credal network specifications and the results of exact inferences on such models. We also deliver a Python tool to easily load CREPO data and interact with CREMA in order to easily evaluate and compare credal network algorithms. Besides the two algorithms already embedded in CREMA, i.e., a credal version of the variable elimination scheme, and an approximate procedure based on a reduction to linear programming \cite{approxlp}, we finally present a very first example of CREPO-based benchmarking in CREMA for a novel heuristic algorithm. The heuristic is applied inside a variable elimination scheme: after any combination step, a geometric reduction is applied to keep a bound on the number of vertices, thus preventing any exponential blow up. This corresponds to an anytime procedure with respect to the value of such bound.

\section{Credal Networks Inference}
The (discrete) variables $\bm{X}:=(X_1,\ldots,X_n)$  of a credal network are in one-to-one correspondence with the nodes of an acyclic directed graph $\mathcal{G}$. Variable $X_i$ takes its values in $\mathcal{X}_i$, $x_i$ being its generic value. Let $\mathrm{Pa}_{X_i}$ denote the \emph{parents} of $X_i$ according to $\mathcal{G}$. In a Bayesian network, a joint mass function $P(\bm{X})$ is obtained from the conditional probability tables $\{P(X_i|\mathrm{Pa}_{X_i})\}_{i=1}^n$. The Markov condition for the directed graph $\mathcal{G}$ induces a number of conditional independence relations leading to the factorization $P(\bm{x})=\prod_{i=1}^n P(x_i|\mathrm{pa}_{X_i})$. A credal network is just a direct generalization of this formalism: each probability mass function $P(X_i|\mathrm{pa}_{X_i})$ is replaced by a (so-called credal) set of probability mass functions. Notation $K(X_i|\mathrm{pa}_{X_i})$ is used. A joint credal set $K(\bm{X})$ is consequently obtained by taking all joint mass functions factorizing as a Bayesian network over the same graph and allowing the conditional probability mass functions of each conditional probability table to take their values from the corresponding credal sets.\footnote{This approach to credal network modelling, based on the notion of \emph{strong independence}, allows for a direct sensitivity analysis interpretation of the results based on these models \cite{cozman2000credal}. The most popular alternative is based on the notion of epistemic irrelevance \cite{debock2017credal}. See \cite{maua2014probabilistic} for a discussion on the relations between the two approaches.} Marginal inference in a credal network is intended as the computation of the lower and upper bounds of the probability of a single variable with respect to the local credal sets, e.g., $\underline{P}(x_q):=\min_{P(X_i|\mathrm{Pa}_{X_i}) \in K(X_i|\mathrm{Pa}_{X_i})} \sum_{\bm{x}\setminus\{x_q\}}\prod_{i=1}^n P(x_i|\mathrm{pa}_{X_i})$. Conditional queries, e.g., $\underline{P}(x_q|x_E)$ are analogously defined. Those queries remains unaffected by the convex closure of the local credal sets \cite{antonucci2008b}. Here we only consider finitely-generated models corresponding to convex sets with a finite number of vertices. As credal networks generalize Bayesian networks, the hardness of those tasks for Bayesian networks \cite{cooper1990}, also concerns credal network inference. Yet, unlike the case of Bayesian networks, the inference remains hard even on simple topologies \cite{maua2014probabilistic}, with the general task belonging to higher complexity classes \cite{de2005inferential}. To better understand this point, consider the classical \emph{variable elimination} (VE) scheme for Bayesian networks \cite{kohlas2012information}, where sums and products involved in the computation of the target are swapped in order to reduce the size of the joint models involved in the sums. For an optimal choice of the (elimination) order in which the sums are performed, the complexity of the procedure is bounded by the exponential of the so-called \emph{treewidth} of graph $\mathcal{G}$ \cite{koller2009}. The same scheme in a credal network require each combination to be performed separately for all the elements of the credal sets associated with the different variables. Although this can be done by considering only the vertices of the credal sets, this might be not sufficient to prevent an exponential blow up. To hinder such growth, the convex hull of any combination can be computed and the inner points removed. Yet, this does not provide theoretical guarantees, thus justifying the hardness results for credal network inference even for bounded treewidth topologies. For this reason approximate schemes have been proposed. One of the most popular is \emph{ApproxLP}, a technique based on a linearization of the optimization required by the inference, that allows to reduce the query to a sequence of linear programming tasks and Bayesian network inferences on a model with the same topology \cite{approxlp}.

\section{The CREPO Repository}\label{sec:crepo}
Let us introduce CREPO, an open and extensible repository of synthetic credal networks, especially designed for the benchmarking of inference algorithms. We first describe its main features and then discuss a number of relevant issues we addressed during its preparation.

\paragraph{Technical features.} 
CREPO is stored at Github to simplify its distribution and versioning.\footnote{\url{https://github.com/IDSIA/crepo}} This is important, as we expect to extend the repository, which now includes only relatively small and simple models ($\leq 10$ nodes, cardinalities between two and three, indegree and number of vertices for credal sets between two and six) in order to make possible to quickly compute exact inferences with the credal VE. CREPO includes $378$ randomly generated networks. The models are specified in UAI-like formats developed for credal networks.\footnote{\url{https://crema-toolbox.readthedocs.io}} The repository also contains the results of exact inferences based on the credal VE and the approximate results achieved with ApproxLP. Moreover, a Python tool\footnote{\url{https://pypi.org/project/crepobenchmark}} was developed to easily access the repository and interact with CREMA as shown in Figure~\ref{fig:code}. 

\begin{figure}[htpb]
%[frame=lines,
%framesep=2mm,
%baselinestretch=0.9,
%tabsize=2,
%{python}
{\footnotesize
\begin{lstlisting}[language=Python]
import crepobenchmark as crb
# Download the inference results
data = crb.get_benchmark_data()
# Save a model from the repository 
crb.save_model("vs_n4_mID2_mD6_mV4_nV2-1",
               "model.uai")
# Run exact inference  
crb.run_crema("model.uai", target=0)
\end{lstlisting}}
\caption{Python code to access CREPO.\label{fig:code}}
\vspace{-18pt}
\end{figure}

\paragraph{Vertices and Linear Constraints.} The local credal sets of a credal network can be assumed to be convex. Thus, we describe a (finitely generated) credal set by explicitly listing their vertices (so-called \emph{V-representation}) or by a finite set of linear constrains on its points (\emph{H-representation}). Algorithms based on credal VE cope with V-representations, as products and sums are achieved separately for all the possible combination of the vertices of different credal sets. Being based on linear programming with respect to local credal sets, optimization-based procedures such as ApproxLP need instead H-representation. Each CREPO network in therefore available in both representations. CREMA supports these formats and offers conversion tools based on polyhedral algorithms \cite{avis2018mplrs}. Yet, when moving from V- to H-representation, numerical issues might arise because of a dimensionality gap: the normalization constraint to be satisfied by its points makes a credal set $K(X)$ an object in the $d$-dimensional space with dimensionality smaller than or equal to $d-1$, where $d:=|\mathcal{X}|$. To address those issues, it suffices to find the actual dimensionality of $K(X)$, i.e., the rank of the matrix containing its vertices, and obtain an orthonormal basis for the vertices, e.g., by Gram-Schmidt orthogonalization \cite{bjorck1994numerics}. The H-representation can be therefore computed with respect to the new basis, where the matrix with the vertices has full rank and hence does not suffer numerical issues, and eventually reformulated in the original basis.

\paragraph{Random Generation of Credal Sets.}
The random credal networks in CREPO have been obtained by extending a Bayesian network tool \cite{ide2004generating}. To generate random credal sets, we stick with the V-representation and a uniform sampling of mass functions from the probability simplex is iterated unless the resulting credal set has the required number of vertices. Dedicated sampling procedure are adopted \cite{smith2004sampling}, as trivial strategies such as sampling uniform numbers separately for each state and then normalizing does not guarantee uniformity. 

\paragraph{Inference Tasks Selection.} When solving marginal or conditional inference tasks in credal networks, an obvious preprocessing task consists in exploiting the d-separation properties of credal networks to simplify the problem. This basically consists in removing the barren nodes, remove the arcs leaving the observed variables, replacing the states of the observed variables with two states only corresponding to the actual observation and its negation, and finally take only the connected component of the model where the queried variable is located. These procedures are embedded in CREMA and the resulting graph is called \emph{requisite} \cite{maua-2020-ijar-b}. In order to have challenging inference tasks, in CREPO we select a marginal and a conditional task maximizing the size of the requisite graph. This roughly corresponds to take leaves as queries for marginal tasks, and roots for conditional ones with observations on the leaves.

\section{Credal VE with  $k$-Reduction}
\begin{figure}[htbp]\label{fig:red}
    \centering
\begin{tikzpicture}[scale=0.4]
    \begin{axis}[
        tick align=outside,
        tick pos=left,
        x grid style={white!69.0196078431373!black},
        xmin=0.0, xmax=1.0,
        %xtick style={color=black},
        %xtick={0.1,0.2,0.3,0.4,0.5,0.6,0.7,0.8,0.9,1,1.1},
        %xticklabels={0.1,0.2,0.3,0.4,0.5,0.6,0.7,0.8,0.9,1.0,1.1},
        %y grid style={white!69.0196078431373!black},
        ymin=0.0, ymax=1.0,%,
        %ytick style={color=black}%,
        %ytick={-0.2,0,0.2,0.4,0.6,0.8,1},
        xticklabels={,,},
        yticklabels={,,}
        ]

        \addplot [semithick, red, mark=*, mark size=3, mark options={solid}, only marks]
        table {%
            0.864536004924652 0.344633071964266
            0.0439424390127909 0.834267041275926
            0.146369596460016 0.562023118426994
            0.0932028139412838 0.674301617404711
            0.700839289168985 0.689625239048539
            0.396247535425765 0.871731181837455
            0.944907220389489 0.614592244388893
            0.393392135810945 0.201281918117575
            0.10095490640833 0.186136091795689
            0.483082749204841 0.0609128873819137
            0.224675572590779 0.369785296748755
            0.482996590168576 0.455149312785737
            0.557659459252472 0.644372823660673
            0.920207113094907 0.581633179323039
            0.44032674656005 0.402856926444781
        };
        \addplot [semithick, blue, mark=*, mark size=3, mark options={solid}, only marks]
        table {%
            0.0439424390127909 0.834267041275926
            0.10095490640833 0.186136091795689
            0.483082749204841 0.0609128873819137
            0.864536004924652 0.344633071964266
            0.944907220389489 0.614592244388893
            0.396247535425765 0.871731181837455
        };
        \addplot [semithick, green!50!black, mark=*, mark size=3, mark options={solid}, only marks]
        table {%
            0.90472161265707 0.479612658176579
        };
        \addplot [semithick, black, dashed]
        table {%
            0.10095490640833 0.186136091795689
            0.0439424390127909 0.834267041275926
        };
        \addplot [semithick, black, dashed]
        table {%
            0.10095490640833 0.186136091795689
            0.483082749204841 0.0609128873819137
        };
        \addplot [semithick, black, dashed]
        table {%
            0.396247535425765 0.871731181837455
            0.0439424390127909 0.834267041275926
        };
        \addplot [semithick, black, dashed]
        table {%
            0.396247535425765 0.871731181837455
            0.944907220389489 0.614592244388893
        };
        \addplot [semithick, black, dashed]
        table {%
            0.864536004924652 0.344633071964266
            0.944907220389489 0.614592244388893
        };
        \addplot [semithick, black, dashed]
        table {%
            0.864536004924652 0.344633071964266
            0.483082749204841 0.0609128873819137
        };
        \addplot [semithick, green!50!black, mark=*, mark size=3, mark options={solid}, only marks]
        table {%
            0.220094987219278 0.852999111556691
        };
        \addplot [semithick, black, dashed]
        table {%
            0.483082749204841 0.0609128873819137
            0.90472161265707 0.479612658176579
        };
        \addplot [semithick, black, dashed]
        table {%
            0.10095490640833 0.186136091795689
            0.0439424390127909 0.834267041275926
        };
        \addplot [semithick, black, dashed]
        table {%
            0.10095490640833 0.186136091795689
            0.483082749204841 0.0609128873819137
        };
        \addplot [semithick, black, dashed]
        table {%
            0.396247535425765 0.871731181837455
            0.0439424390127909 0.834267041275926
        };
        \addplot [semithick, black, dashed]
        table {%
            0.396247535425765 0.871731181837455
            0.90472161265707 0.479612658176579
        };
        \addplot [semithick, black]
        table {%
            0.220094987219278 0.852999111556691
            0.10095490640833 0.186136091795689
        };
        \addplot [semithick, black]
        table {%
            0.220094987219278 0.852999111556691
            0.90472161265707 0.479612658176579
        };
        \addplot [semithick, black]
        table {%
            0.483082749204841 0.0609128873819137
            0.10095490640833 0.186136091795689
        };
        \addplot [semithick, black]
        table {%
            0.483082749204841 0.0609128873819137
            0.90472161265707 0.479612658176579
        };
    \end{axis}
\end{tikzpicture}
%{\input{reduction.tex}
\caption{A $4$-reduction: inner points of the original convex set are red, its vertices blue, midpoints green.}
\vspace{-18pt}
\end{figure}
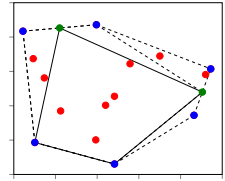

Let us call \emph{reduction} any transformation of a convex set reducing its number of points and achieved by removing or replacing other points. Convex hull, intended here as the removal of the inner points after the convex closure, is a reduction not providing guarantees on the size of its output (apart from the trivial case of credal sets over Boolean variables). Although typically faster than convex hull, discarding Pareto-dominated points might encounter the same drawback \cite{maua2012updating}. This is the case also for Pareto-dominance relaxations  \cite{marinescu2017multi}. In practice, good approximations can be obtained in this way, but it is not possible to predict in advance the minimum relaxation level keeping the running time under some threshold. To address such limitations, we propose here a $k$-reduction, intended as a reduction providing an upper bound on the number of points of the output. Sampling (or picking) $k$ points from a credal set or from its vertex is an obvious $k$-reduction, while a more sophisticated approach is depicted in Figure \ref{fig:red} and defined as follows.
\begin{definition}[$k$-reduction]\label{def:red}
Given convex set $K$ in input, take its vertices $e(K)$. Find the two distinct points $p',p''$ of $e(K)$ at minimum distance. Remove $p'$ and $p''$ from $e(K)$ and return in output $e(K) \cup \{(p'+p'')/2\}$. Repeat this procedure until the number of points becomes equal to $k$.
\end{definition}

Credal VE might take exponential time even with bounded-treewidth models because of an unbounded growth in the number of points involved in the combinations. Taking the convex hull after any combination might be not enough, while adding the above $k$-reduction prevents such growth. As convex hull algorithms might be slow because of the high number of input points and/or space dimensionality \cite{barber1996quickhull}, it is important to note that the procedure in Definition \ref{def:red} does not require the convex hull to be executed again after the reductions, as the midpoint of the two vertices removed from the set can be proved to be a vertex of the new set (see supplementary material). Moreover, those midpoints are inner points of the original convex set, thus the algorithm provides an inner approximation of the exact bounds of the query. The performance of the combination of such method with credal VE for two different threshold levels and Kullback-Leibler as (pseudo) distance in the reduction has been tested in CREMA on the CREPO tasks. Table \ref{tab:exp} shows that the new method can nicely trade off accuracy and execution time (the \emph{speed-up} is the ratio between execution time of credal VE and that of the algorithm).

\begin{table}[htbp]
\floatconts{tab:exp}%
{\caption{Benchmarking $k$-reduction in CREPO.}}%
{
\begin{tabular}{lrr}
  \toprule
  \bfseries Method & \bfseries RMSE & \bfseries Speed up\\
  \midrule
  ApproxLP & 0.0058 & 32.107 \\
  $10$-reduction + CVE & 0.0011 & 86.849 \\
  $5$-reduction + CVE & 0.0045 & 147.066 \\
  \bottomrule
\end{tabular}}
\end{table}

\section{Conclusion and Outlooks}
A first repository to benchmark credal network inference has been presented together with a novel scheme to speed up credal variable elimination. As a future work, we intend to expand CREPO with new and more challenging inference tasks, and add to CREMA other algorithms, such as those in 
\cite{antonucci2010generalized,de2004inference}, to promote a deeper understanding of the existing algorithms and the design of new schemes. For the $k$-reduction, we want to achieve a further validation and investigate the analogies between our geometrical approach and the information-theoretic method proposed in \cite{cano2002using}.

\clearpage
\appendix
\section{Supplementary Material}
Consider a closed and convex set $C$ in $\mathbb{R}^d$. Let $d$ denote also the dimension of $C$. The vertices of $C$ are assumed to be finite and denoted as $e(C)$. A \emph{hyperplane} $H$ in $\mathbb{R}^d$ can be parametrized by a pair $(\bm{v},w)$, with $\bm{v}\in\mathbb{R}^d$ and $w\in\mathbb{R}$ as follows:
\begin{equation}
H_{\bm{v},w} := \{ \bm{x} \in \mathbb{R}^d : \bm{v} \cdot \bm{x} = w \}\,.
\end{equation}
The \emph{segment} $S$ connecting points $\bm{a},\bm{b} \in \mathbb{R}^d$ is instead:
\begin{equation}
S_{\bm{a},\bm{b}} := \{ \bm{x} \in \mathbb{R}^d := \lambda \bm{a} + (1-\lambda) \bm{b}, 0 \leq \lambda \leq 1 \}\,.
\end{equation}
\begin{definition}
$H_{\bm{v},w}$ is a \emph{supporting} hyperplane for $C$ passing through $\bm{x}^*\in C$ if and only if $\bm{x}^* \in P_{\bm{v},w}$ and $\bm{v}\cdot\bm{x}\leq \bm{v}\cdot\bm{x}^*$ for each $\bm{x}\in C$.
\end{definition}
\begin{definition}
A point $\bm{x}^*$ belongs to the \emph{boundary} $b(C)$ of the convex set $C$ if and only if there is at least a supporting hyperplane for $C$ passing through $\bm{x}^*$.
\end{definition}

We use notation $\mathrm{CH}$ for the convex hull of a set of points, e.g., $C:=\mathrm{CH}[e(C)]$. The following result holds.

\begin{lemma}
Let $C$ be a convex set in $\mathbb{R}^d$ 
such that $d$ is also its dimension (i.e., 
Given $\bm{a},\bm{b}\in e(C)$, let $\bm{x}^*:=\frac{1}{2}(\bm{a}+\bm{b})$. Let also $C':=\mathrm{CH}[e(C)\setminus\{\bm{a},\bm{b}\}]$, while $H_{\bm{v},w}$ denotes a supporting hyperplane for $C$ through $\bm{x}^*$. It holds that, if $S_{\bm{a},\bm{b}} \subset H_{\bm{v},w}$, then $\bm{x}^* \not \in C'$.
\begin{proof}
By construction $\bm{x} \in S_{\bm{a},\bm{b}}$. Assume, \emph{ad absurdum}, $\bm{x}^* \in C'$. Thus, $\bm{x}^*$ should be a convex combination of the vertices of $C'$, i.e.
\begin{equation}
\bm{x}^* = \sum_{\bm{z}\in e(C)\setminus \{\bm{a},\bm{b}\}} \lambda_{\bm{z}} \bm{z}\,,
\end{equation}
where $\lambda_{\bm{z}} \geq 0$ for each $\bm{z}\in e(C)\setminus\{\bm{a},\bm{b}\}$ and $\sum_{\bm{z}\in e(C)\setminus\{\bm{a},\bm{b}\}} \lambda_{\bm{z}}=1$. Take the scalar product by $\bm{v}$:
\begin{equation}
\bm{v}\cdot\bm{x}^*=
\bm{v}\cdot \left[ \sum_{\bm{z}\in e(C)\setminus\{\bm{a},\bm{b}\}} \lambda_{\bm{z}} \bm{z} \right]
=\sum_{\bm{z}\in e(C)\setminus\{\bm{a},\bm{b}\}} \lambda_{\bm{z}} \bm{v}\cdot\bm{z}\,.
\end{equation}
By supporting hyperplane definition and simple algebra:
\begin{equation}
\bm{v}\cdot\bm{x}^*=\sum_{\bm{z}\in e(C)\setminus\{\bm{a},\bm{b}\}} \lambda_{\bm{z}} \bm{v}\cdot\bm{z}\leq\sum_{\bm{z}\in e(C)\setminus\{\bm{a},\bm{b}\}} \lambda_{\bm{z}} \bm{v}\cdot\bm{x}^*=\bm{v}\cdot\bm{x}^*\,.
\end{equation}
This implies $\bm{z}\in H_{\bm{v},w}$ for each $z \in e(C)\setminus\{\bm{a},\bm{b}\}$. As also $\bm{a}$ and $\bm{b}$ belong to $H_{\bm{v},w}$, we have $e(C) \subset H_{\bm{v},w}$. In other words $C$ is included in a hyperplane and it coincides with its boundary, but this is against the original assumption about the dimension of $C$.
\end{proof}
\end{lemma}
As a consequence of this lemma we have that, in Definition \ref{def:red}, the midpoint of the two vertices added to $C$ is a vertex of the new set. This simply following that the two points at minimum (Euclidean) distance belong to a same edge of a convex polytope and the credal set can be always parametrized in order to have full dimension (see discussion in Section \ref{sec:crepo}. When coping with non-Euclidean distances, to have the same result, the two points at minimum distance in Definition \ref{def:red} should be detected with the additional condition of belonging to a same edge. 

\begin{thebibliography}{28}
    \providecommand{\natexlab}[1]{#1}
    \providecommand{\url}[1]{\texttt{#1}}
    \expandafter\ifx\csname urlstyle\endcsname\relax
    \providecommand{\doi}[1]{doi: #1}\else
    \providecommand{\doi}{doi: \begingroup \urlstyle{rm}\Url}\fi
    
    \bibitem[Antonucci and Zaffalon(2008)]{antonucci2008b}
    Alessandro Antonucci and Marco Zaffalon.
    \newblock Decision-theoretic specification of credal networks: A unified
    language for uncertain modeling with sets of {B}ayesian networks.
    \newblock \emph{International Journal of Approximate Reasoning}, 49\penalty0
    (2):\penalty0 345--361, 2008.
    
    \bibitem[Antonucci et~al.(2004)Antonucci, Salvetti, and
    Zaffalon]{antonucci2004a}
    Alessandro Antonucci, Andrea Salvetti, and Marco Zaffalon.
    \newblock Hazard assessment of debris flows by credal networks.
    \newblock In C.~Pahl-Wostl, S.~Schmidt, A.~E. Rizzoli, and A.~J. Jakeman,
    editors, \emph{iEMSs 2004: Complexity and Integrated Resources Management,
        Transactions of the 2nd Biennial Meeting of the International Environmental
        Modelling and Software Society}, pages 98--103. iEMSs, 2004.
    
    \bibitem[Antonucci et~al.(2009)Antonucci, Br{\"u}hlmann, Piatti, and
    Zaffalon]{antonucci2009credal}
    Alessandro Antonucci, Ralph Br{\"u}hlmann, Alberto Piatti, and Marco Zaffalon.
    \newblock Credal networks for military identification problems.
    \newblock \emph{International Journal of Approximate Reasoning}, 50\penalty0
    (4):\penalty0 666--679, 2009.
    
    \bibitem[Antonucci et~al.(2010)Antonucci, Sun, De~Campos, and
    Zaffalon]{antonucci2010generalized}
    Alessandro Antonucci, Yi~Sun, Cassio~P. De~Campos, and Marco Zaffalon.
    \newblock Generalized loopy {2U}: a new algorithm for approximate inference in
    credal networks.
    \newblock \emph{International Journal of Approximate Reasoning}, 51\penalty0
    (5):\penalty0 474--484, 2010.
    
    \bibitem[Antonucci et~al.(2015)Antonucci, de~Campos, Huber, and
    Zaffalon]{approxlp}
    Alessandro Antonucci, Cassio~P. de~Campos, David Huber, and Marco Zaffalon.
    \newblock Approximate credal network updating by linear programming with
    applications to decision making.
    \newblock \emph{International Journal of Approximate Reasoning}, 58:\penalty0
    25--38, 2015.
    
    \bibitem[Avis and Jordan(2018)]{avis2018mplrs}
    David Avis and Charles Jordan.
    \newblock Mplrs: A scalable parallel vertex/facet enumeration code.
    \newblock \emph{Mathematical Programming Computation}, 10\penalty0
    (2):\penalty0 267--302, 2018.
    
    \bibitem[Barber et~al.(1996)Barber, Dobkin, and Huhdanpaa]{barber1996quickhull}
    C.~Bradford Barber, David~P. Dobkin, and Hannu Huhdanpaa.
    \newblock The quickhull algorithm for convex hulls.
    \newblock \emph{ACM Transactions on Mathematical Software (TOMS)}, 22\penalty0
    (4):\penalty0 469--483, 1996.
    
    \bibitem[Bj{\"o}rck(1994)]{bjorck1994numerics}
    {\AA}ke Bj{\"o}rck.
    \newblock Numerics of {G}ram-{S}chmidt orthogonalization.
    \newblock \emph{Linear Algebra and Its Applications}, 197:\penalty0 297--316,
    1994.
    
    \bibitem[Borgelt et~al.(2000)Borgelt, Gebhardt, and
    Kruse]{borgelt2000possibilistic}
    Christian Borgelt, J{\"o}rg Gebhardt, and Rudolf Kruse.
    \newblock Possibilistic graphical models.
    \newblock In \emph{Computational Intelligence in Data Mining}, pages 51--67.
    Springer, 2000.
    
    \bibitem[Cano and Moral(2002)]{cano2002using}
    Andr{\'e}s Cano and Serafin Moral.
    \newblock Using probability trees to compute marginals with imprecise
    probabilities.
    \newblock \emph{International Journal of Approximate Reasoning}, 29\penalty0
    (1):\penalty0 1--46, 2002.
    
    \bibitem[Cooper(1990)]{cooper1990}
    Gregory~F. Cooper.
    \newblock The computational complexity of probabilistic inference using
    {B}ayesian belief networks.
    \newblock \emph{Artificial Intelligence}, 42:\penalty0 393--405, 1990.
    
    \bibitem[Cozman(2000)]{cozman2000credal}
    Fabio~G Cozman.
    \newblock Credal networks.
    \newblock \emph{Artificial intelligence}, 120\penalty0 (2):\penalty0 199--233,
    2000.
    
    \bibitem[De~Bock(2017)]{debock2017credal}
    Jasper De~Bock.
    \newblock Credal networks under epistemic irrelevance.
    \newblock \emph{International Journal of Approximate Reasoning}, 85:\penalty0
    107--138, 2017.
    
    \bibitem[de~Campos and Cozman(2004)]{de2004inference}
    Cassio~P. de~Campos and Fabio~G. Cozman.
    \newblock Inference in credal networks using multilinear programming.
    \newblock In \emph{Proceedings of the Second Starting AI Researcher Symposium},
    pages 50--61. Citeseer, 2004.
    
    \bibitem[De~Campos and Cozman(2005)]{de2005inferential}
    Cassio~P. De~Campos and Fabio~G. Cozman.
    \newblock The inferential complexity of bayesian and credal networks.
    \newblock In \emph{IJCAI}, volume~5, pages 1313--1318. Citeseer, 2005.
    
    \bibitem[Estrada-Lugo et~al.(2019)Estrada-Lugo, De~Angelis, and
    Patelli]{estrada2019probabilistic}
    Hector~Diego Estrada-Lugo, Marco De~Angelis, and Edoardo Patelli.
    \newblock Probabilistic risk assessment of fire occurrence in residential
    buildings: Application to the grenfell tower.
    \newblock In \emph{13th International Conference on Applications of Statistics
        and Probability in Civil Engineering}, 2019.
    
    \bibitem[Huber et~al.(2020)Huber, Cabañas, Antonucci, and
    Zaffalon]{huber2020a}
    David Huber, Rafael Cabañas, Alessandro Antonucci, and Marco Zaffalon.
    \newblock {CREMA}: A {J}ava library for credal network inference.
    \newblock In Manfred Jaeger and Thomas~Dyhre Nielsen, editors,
    \emph{Proceedings of the 10th International Conference on Probabilistic
        Graphical Models ({PGM} 2020)}, Proceedings of Machine Learning Research,
    Aalborg, Denmark, 2020. PMLR.
    
    \bibitem[Ide et~al.(2004)Ide, Cozman, and Ramos]{ide2004generating}
    Jaime~Shinsuke Ide, F{\'a}bio~G. Cozman, and Fabio~T. Ramos.
    \newblock Generating random {B}ayesian networks with constraints on induced
    width.
    \newblock In \emph{ECAI}, volume~16, page 323, 2004.
    
    \bibitem[Kohlas(2012)]{kohlas2012information}
    J{\"u}rg Kohlas.
    \newblock \emph{Information algebras: Generic structures for inference}.
    \newblock Springer Science \& Business Media, 2012.
    
    \bibitem[Koller and Friedman(2009)]{koller2009}
    Daphne Koller and Nir Friedman.
    \newblock \emph{Probabilistic graphical models: principles and techniques}.
    \newblock MIT, 2009.
    
    \bibitem[Marinescu et~al.(2017)Marinescu, Razak, and
    Wilson]{marinescu2017multi}
    Radu Marinescu, Abdul Razak, and Nic Wilson.
    \newblock Multi-objective influence diagrams with possibly optimal policies.
    \newblock In \emph{Proceedings of the AAAI Conference on Artificial
        Intelligence}, volume~31, 2017.
    
    \bibitem[Mau{\'a} et~al.(2012)Mau{\'a}, De~Campos, and
    Zaffalon]{maua2012updating}
    Denis~D Mau{\'a}, Cassio~P De~Campos, and Marco Zaffalon.
    \newblock Updating credal networks is approximable in polynomial time.
    \newblock \emph{International Journal of Approximate Reasoning}, 53\penalty0
    (8):\penalty0 1183--1199, 2012.
    
    \bibitem[Mau{\'a} et~al.(2014)Mau{\'a}, De~Campos, Benavoli, and
    Antonucci]{maua2014probabilistic}
    Denis~D. Mau{\'a}, Cassio~P. De~Campos, Alessio Benavoli, and Alessandro
    Antonucci.
    \newblock Probabilistic inference in credal networks: new complexity results.
    \newblock \emph{Journal of Artificial Intelligence Research}, 50:\penalty0
    603--637, 2014.
    
    \bibitem[Mauá and Cozman(2020)]{maua-2020-ijar-b}
    Denis~D. Mauá and Fabio~G. Cozman.
    \newblock Thirty years of credal networks: Specification, algorithms and
    complexity.
    \newblock \emph{International Journal of Approximate Reasoning}, 126:\penalty0
    133--157, 2020.
    
    \bibitem[Pearl(1988)]{pearl2014probabilistic}
    Judea Pearl.
    \newblock \emph{Probabilistic reasoning in intelligent systems: networks of
        plausible inference}.
    \newblock Morgan Kaufmann Publishers Inc., San Francisco, CA, USA, 1988.
    
    \bibitem[Smith and Tromble(2004)]{smith2004sampling}
    Noah~A. Smith and Roy~W. Tromble.
    \newblock Sampling uniformly from the unit simplex.
    \newblock \emph{Johns Hopkins University, Technical Report}, 29, 2004.
    
    \bibitem[Yaghlane and Mellouli(2008)]{yaghlane2008inference}
    Boutheina~Ben Yaghlane and Khaled Mellouli.
    \newblock Inference in directed evidential networks based on the transferable
    belief model.
    \newblock \emph{International Journal of Approximate Reasoning}, 48\penalty0
    (2):\penalty0 399--418, 2008.
    
    \bibitem[Zaffalon et~al.(2020)Zaffalon, Antonucci, and Cabañas]{zaffalon2020a}
    Marco Zaffalon, Alessandro Antonucci, and Rafael Cabañas.
    \newblock Structural causal models are (solvable by) credal networks.
    \newblock In Manfred Jaeger and Thomas~Dyhre Nielsen, editors,
    \emph{Proceedings of the 10th International Conference on Probabilistic
        Graphical Models ({PGM} 2020)}, Proceedings of Machine Learning Research.
    PMLR, September 2020.
    
\end{thebibliography}
\end{document}